\documentclass[paper]{article}
\usepackage{graphicx,xcolor}
\usepackage{float}
\usepackage[fleqn]{amsmath}
\usepackage{newtxtext}
\usepackage[varg]{newtxmath}

\usepackage{color}

\title{In-context learning of solution to simple linear regression task using
transformer with masked linear self-attention}

\usepackage{url}
\urldef{\mailsa}\path|hagi@edu.mie-u.ac.jp|

\author{Katsuyuki Hagiwara}
\date{Faculty of Education, Mie University\\hagi@edu.mie-u.ac.jp}

\def\q{{\boldsymbol{q}}}

\def\A{{\mathbf A}}
\def\O{{\mathbf O}}
\def\P{{\mathbf P}}
\def\Q{{\mathbf Q}}
\def\W{{\mathbf W}}
\def\H{{\mathbf H}}
\def\oH{\overline{\mathbf H}}

\def\LSA{{\rm LSA}}
\def\ox{{\overline{x}}}
\def\oy{{\overline{y}}}
\def\vTheta{{\boldsymbol{\Theta}}}

\def\ey{{\widehat{y}}}
\def\etheta{{\widehat{\theta}}}

\def\Win{{\mathbf W}_{\rm in}}
\def\Wout{{\mathbf W}_{\rm out}}
\def\oWout{\overline{\mathbf W}_{\rm out}}
\def\bin{{\boldsymbol{b}_{\rm in}}}
\def\bout{b_{\rm out}}
\def\TF{{\rm TF}}
\def\D{{\mathbf D}}

\def\TFout{Q_{\rm out}}
\def\M{{\mathbf M}}
\def\oM{\overline{\mathbf M}}
\def\I{{\mathbf I}}

\begin{document}
\maketitle
\begin{abstract}
In-context learning is a remarkable property of transformers and has
recently received a lot of interest. In many studies of in-context
learning, it has been shown that transformers are capable of
implementing solvers for linear and non-linear regression problems, in
which the most of them implement gradient descent algorithm. However, it
is still unclear whether those implementations have actually been
acquired through training. In this paper, we constructed a transformer
with masked linear self-attention, which in-context learns the least
squares estimate in a simple regression task. The point here is that the
closed form (analytical) solution is obtained by using layer
normalization rather than an approximate solution based on the gradient
descent algorithm.  Then, we showed an experimental example, in which,
by taking the non-uniqueness of the transformer representation into
account, our implementation for a closed form solution is used in the
transformer trained with the $\ell_1$ regularization when the target
output is the least squares estimate.

{\bf keywords} in-context learning, transformer, linear self-attention,
a simple linear regression, layer normalization, mask

\end{abstract}

\section{Introduction}

In-context learning is a remarkable property of transformers, which form
the basis of large language models such as GPT-3 \cite{Brown2020}, and
it has been the subject of recent research. In-context learning is a
capability of transformers to learn new tasks at inference time solely
by processing context (sequence) provided in the prompt, without
modifying the model's underlying weights or undergoing additional
training.  It is natural to think that transformers acquire, through
training, the algorithms to solve tasks via in-context learning. In this
regard, it has been demonstrated that transformers are capable of
implementing various algorithms, particularly in regression tasks
\cite{Ahn2023,Akyurek2022,Bai2023,Garg2023,Oswald2022}.

\cite{Garg2023} empirically studied the in-context learning abilities of
transformers for various function classes in machine learning, including
the linear function class. In particular, for linear functions, a
trained transformer performs similarly to the least squares
solution. \cite{Oswald2022} provided an explicit construction of a
single linear self-attention layer that implements a single step of the
gradient descent algorithm on the mean squared error loss. Additionally,
they empirically showed that several self-attention layers can
iteratively perform curvature correction, improving upon the plain
gradient descent algorithm. \cite{Akyurek2022} proved that a transformer
can implement a gradient descent algorithm and a single Sherman-Morrison
update for a ridge regression problem. While the works of
\cite{Garg2023,Oswald2022,Akyurek2022} do not consider the training
phase, \cite{Ahn2023} proved the global minimum of the training
objective implements a single iteration of preconditioned gradient
descent when the target is a linear function. In \cite{Ahn2023}, even
though the training objective is the expected error, the results of the
numerical experiments are shown to closely match the theory. The similar
result has obtained also in \cite{Mahankali2023}.  More recently,
\cite{Bai2023} showed that transformers can implement a broad class of
standard machine learning algorithms in context, such as least squares,
ridge regression, and Lasso. In contrast to \cite{Akyurek2022},
\cite{Bai2023} precisely evaluated the prediction performance in terms
of network size, and shown a near-optimal predictive
power. \cite{Bai2023} also demonstrated the algorithm selection ability
of transformers, such as regularization selection according to
validation error for ridge regression.

These works have investigated regression problems, finding that the
solver implemented by a transformer is basically the gradient descent
algorithm except for a part of \cite{Akyurek2022}. This is natural since
the attention mechanism in a transformer computes the product of the
inputs, which is exactly what is required for the gradient descent
method for especially linear regression problems. In a linear regression
problem, however, the least squares estimate is analytically obtained by
calculating the matrix inverse, which requires division. Consequently,
the gradient descent algorithm implicitly calculates this division by
repeatedly performing multiplication and addition. On the other hand,
\cite{Akyurek2022} showed that a transformer can implement a single
Sherman-Morrison update for a ridge regression problem, in which the
division is implemented using layer normalization. Applying this idea,
\cite{Hagiwara2026} showed that a transformer with linear self-attention
can approximately implement a closed form solution to a simple
regression task. Although it is found that the implementation is mainly
used in the transformer trained with $\ell_1$ regularization when the
target output is the least squares estimate, unfortunately, the trained
transformer is not fully explained. This may be caused by the assumption
in the implementation, which is that the sequence length in the prompt
is sufficiently large. In this paper, we provide a construction of a
transformer with a masked linear-self attention, which implements a
closed form (analytical) solution to the least squares estimate of a
simple regression task. The implementation does not require the above
assumption in \cite{Hagiwara2026} due to the introduction of a mask.  We
then present a numerical example, in which this implementation is
actually used in the transformer trained with the $\ell_1$
regularization when the target is the least squares estimate.

This paper is organized as follows. Section 2 formulates our problem
setting. Section 3 presents the construction of a transformer that
calculates the least squares estimate based on in-context
samples. Section 4 provides a numerical example of training to
demonstrate that this construction is actually valid. Finally,
Section 5 concludes the paper and discusses future work.

\section{Problem setting}

\subsection{Notations}

In this paper, $\O_{I,J}$ is the $I\times J$ zero matrix and $\I_{J}$ is
a $J\times J$ identity matrix. Note that we use this notation also for
vectors, where either $I$ or $J$ equals $1$.  For an $I\times J$ matrix
$\A$, $\A[i,j]$ is the $(i,j)$-entry of $\A$, $\A[i,:]$ is the $i$-th
row vector of $\A$ and $\A[:,j]$ is the $j$-th column vector of $\A$ for
$i=1,\ldots,I$ and $j=1,\ldots,J$. When $\A$ is an $I\times 1$ vector,
$\A[i]$ denotes its $i$-th entry.
$\A[i_1:i_2,:]$ is a submatrix of $\A$ consisting of rows $i_1$ through $i_2$.
The same applies to submatrices consisting of columns.

\subsection{Setting of in-context learning of transformer}

We next explain the transformer-based in-context learning of a simple
linear regression task. Let $M$ be the number of training data for the
transformer. Let $(x,y)$ be a pair of input-output variables in a
simple regression problem.  At each $m$ in $\{1,\ldots,M\}$,
a set of $N$ random samples of $(x,y)$, which is denoted by
$\{(x_{m,n},y_{m,n}):n=1,\ldots,N\}$, are generated.
We denote the $m$-th prompt (input to the transformer)
by $\P_m$, which is an $(N+1)\times 3$ matrix and whose $n$-th row is
\begin{align}
\label{eq:slr-prompt}
\P_m[n,:]=
\begin{bmatrix}
1 & x_{m,n} & y_{m,n}
\end{bmatrix}, 
\end{align}
where we define $y_{m,N+1}:=0$ and $x_{m,N+1}:=u_m$ that is a prediction
point; e.g. see \cite{Akyurek2022}. Thus, the input sequence length is $N+1$
and the number of in-context samples is $N$. We assume that, for each
$x_{m,n}$, $y_{m,n}$ is generated by
\begin{align}
y_{m,n}=\theta_{m,0}+\theta_{m,1}x_{m,n}+\varepsilon_{m,n}, 
\end{align}
where $\varepsilon_{m,1},\ldots,\varepsilon_{m,N}$, $m=1,\ldots,M$ are
i.i.d. additive noises from a probability distribution with mean $0$ and
variance $\sigma^2<\infty$. Our goal is to obtain the least squares
prediction at $x=u_m$ for the prompt $\P_m$.  Therefore, for each $m$, we need to
calculate the least squares solution using
\begin{align}
\label{eq:D_m}
\D_m=\{(x_{m,n},y_{m,n}):n=1,\ldots,N\},
\end{align}
which is the set of in-context samples. The important point is that the
regression lines can be different for each $m$. Thus, we may assume that
$(\theta_{m,0},\theta_{m,1})$ are sampled from a probability
distribution for each $m$. However, in the construction of the
transformer, we do not make any specific assumptions on the underlying
probability distribution of $(\theta_{m,0},\theta_{m,1})$. We
also do not make any specific assumptions on the underlying probability
distribution of $x_{m,n}$ and $\varepsilon_{m,n}$.

\subsection{Training data for the transformer}

For the $m$-th training data $\D_m$ defined in (\ref{eq:D_m}), we define
\begin{align}
\ox_m&:=\frac{1}{N}\sum_{n=1}^Nx_{m,n}\\ 
\oy_m&:=\frac{1}{N}\sum_{n=1}^Ny_{m,n}\\ 
\label{eq:Vm}
V_m&:=\frac{1}{N}\sum_{n=1}^N(x_{m,n}-\ox_m)^2\\ 
\label{eq:Cm}
C_m&:=\frac{1}{N}\sum_{n=1}^N(x_{m,n}-\ox_m)(y_{m,n}-\oy_m).
\end{align}

For a simple regression problem, it is easy to see that the prediction at
$x=u_m$ using the least squares solution based on $\D_m$ is given by
\begin{align}
\label{eq:lse-estimate-at-u_m}
\ey_m(u_m):=\oy_m+\frac{C_m}{V_m}(u_m-\ox_m)
\end{align}
(see e.g. \cite{RD2002}).

Below, we construct the transformer that receives $\P_m$ as an input and 
outputs $\ey_m(u_m)$ for $m=1,\ldots,M$.
Thus, the transformer calculates the least squares estimate at
the prediction point $u_m$ using in-context samples $\D_m$ at each $m$.

\subsection{Linear self-attention}

We define a linear self-attention (LSA) with a partial mask.
It  receives an $(N+1)\times
D$ matrix $\Q$ as an input and outputs $(N+1)\times
D$ matrix defined by
\begin{align}
\label{eq:LSA}
\LSA_{\vTheta}(\Q)&:=(\Q\W_3)(\M\Q\W_1^\top)^\top(\M\Q\W_2)\notag\\
&=(\Q\W_3)(\W_1(\M\Q)^\top(\M\Q)\W_2),
\end{align}
where $\vTheta=\{\W_1,\W_2,\W_3\}$ is an ordered set of parameters, in which
$\W_1$, $\W_2$ and $\W_3$ are $D\times D$ matrices, and
$\M$ is an $(N+1)\times(N+1)$ matrix which is defined by
\begin{align}
\M:=\begin{bmatrix}
\I_N & \O_{N,1}\\
\O_{1,N}  & 0
   \end{bmatrix}.
\end{align}
Here, we have
\begin{align}
\M\Q=\oM\odot\Q, 
\end{align}
where $\odot$ is the Hadamard product and $\oM$ is an $(N+1)\times D$
matrix which is defined by
\begin{align}
\oM[i,j]:=\begin{cases}
          1 & i\neq N+1\\
0\ & i=N+1
         \end{cases}. 
\end{align}
$\oM$ is known as a mask. As a result, 
\begin{align}
\M\Q=\begin{cases}
      \Q[i,j] & i\neq N+1\\
      0 & i=N+1\\
     \end{cases}. 
\end{align}
Therefore, the last entries of the sequence in $\Q$ are set to zero by
this mask. This is necessary to distinguish between the members of the
sequence in the calculation. The details will be provided later.
Although the above LSA partially uses the mask, in what follows, this
LSA will be referred to simply as LSA. The operation using LSA without
masking is precisely discussed in \cite{Hagiwara2025} and the insight in
\cite{Hagiwara2025} is also used in this paper.

\subsection{Transformer structure}

\begin{figure}[ht]
\centering
\begin{minipage}{70mm}
\hspace{14mm}\centering
\includegraphics[width=40mm]{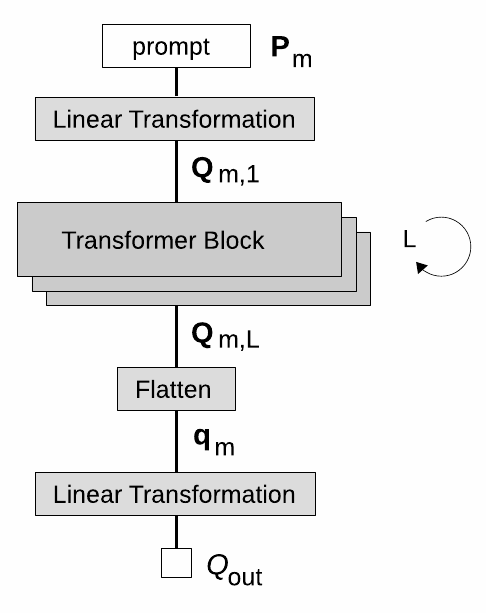}\\
(a) Main stream
\end{minipage}

\vspace{2mm}

\begin{minipage}{70mm}
\centering
\includegraphics[width=38mm]{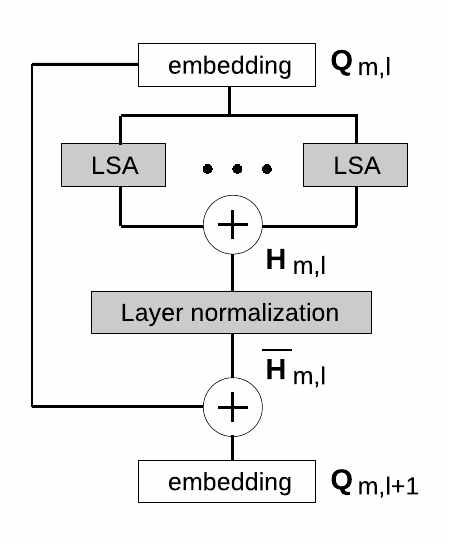}\\
(b) Transformer block
\end{minipage}

\caption{Structure of transformer}
\label{fig:structure}
\end{figure}

The transformer structure considered in this paper is illustrated in
Fig. \ref{fig:structure}, in which grey blocks are operation blocks. The
transformer here consists of input linear transformation, stacking
transformer blocks, flatten and output linear transformation
(Fig. \ref{fig:structure} (a)).  The transformer block consists of
multi-head LSA (MHLSA) block with layer normalization followed by a skip
connection (Fig. \ref{fig:structure} (b)).  It receives the $m$-th
prompt $\P_m$ defined in (\ref{eq:slr-prompt}) and, for simplicity,
outputs a scalar value whose target is $\ey_m(u_m)$ given by
(\ref{eq:lse-estimate-at-u_m}).

Let $\Win$ and $\bin$ be a $D\times 3$ input weight matrix and $D\times 1$
input bias vector respectively. 
The embedding of the prompt $\P_m$ 
by an affine transformation with $(\Win,\bin)$ is denoted by $\Q_{m,1}$ whose 
size is $(N+1)\times D$. More precisely, we define
\begin{align}
\label{eq:input-embedding}
\Q_{m,1}[n,:]^\top:=\Win\P_m[n,:]^\top+\bin
\end{align}
for $n=1,\ldots,N+1$.  The number of layers is denoted by $L$.  For the
$m$-th training data, the input to the $l$-th layer is denoted by
$\Q_{m,l}$, $l=1,\ldots,L$.  Thus, the above $\Q_{m,1}$ is the input to
the first layer.  

The parameter of the LSA block of the $k$-th head in the $l$-th layer is
denoted by $\vTheta_{l,k}=\{\W_{l,k,1},\W_{l,k,2},\W_{l,k,3}\}$. The
output of the MHLSA in the $l$-th layer for $m$-th training data is
\begin{align}
\label{eq:MHLSAoutput}
\H_{m,l}:=\sum_{k=1}^{K}\H_{m,l,k}\\
\label{eq:LSAoutput}
\H_{m,l,k}:=\LSA_{\vTheta_{l,k}}(\Q_{m,l}),
\end{align}
where $K$ is the number of heads. 
The size of $\H_{m,l}$ is $(N+1)\times D$. The number of heads is common for all
layers. We employ the layer normalization along the model dimension (embedding).
We define
\begin{align}
\label{eq:LN-mu}
\mu_{m,l,n}&:=\frac{1}{D}\sum_{d=1}^D\H_{m,l}[n,d]\\
\label{eq:LN-sigma2}
\sigma^2_{m,l,n}&:=\frac{1}{D}\sum_{d=1}^D
(\H_{m,l}[n,d]-\mu_{m,l,n})^2.
\end{align}
Then, the output of the layer normalization with these parameters is
denoted by $\oH_{m,l}$ whose $(n,d)$-entry is 
\begin{align}
\label{eq:after-LN}
\oH_{m,l}[n,d]:=(\H_{m,l}[n,d]-\mu_{m,l,n})/\sigma_{m,l,n}
\end{align}
for $d=1,\ldots,D$ at each $n$.  Finally, the output of the $l$-th layer
is given by
\begin{align}
\label{eq:after-skip-connection}
\Q_{m,l+1}=\Q_{m,l}+\oH_{m,l},
\end{align}
where the first term of the right-hand side comes from a skip
connection. Note that, in our model, the layer normalization is applied
before the skip connection.

$\Q_{m,L}$ is the output of the last layer. It is then flattened and
obtain $\q_m$ which is a $D(N+1)\times 1$ vector.  Let $\Wout$ and
$\bout$ be a $D(N+1)\times 1$ output weight matrix and output bias respectively.
We then obtain the transformer output by
\begin{align}
\label{eq:transformer-output}
\TFout(\P_m):=\Wout^\top\q_m+\bout
\end{align}
which is a scalar value.

We refer to this transformer as $\TF(D,L,K)$, where $D$, $L$ and $K$ are
the model dimension, the number of layers and heads respectively. 

\section{A construction of transformer}
\label{sec:construction-of-transformer}

We here gives a construction of the transformer that calculates the
least squares estimate at the prediction point. We here fix the
structure of the transformer, in which the model dimension $D=4$, the
number of layers $L=2$ and the number of heads $K=2$.  Therefore, the
transformer is $\TF(4,2,2)$.  Now, the $m$-th prompt to $\TF(4,2,2)$ is
$\P_m$ defined in (\ref{eq:slr-prompt}).

\subsection{Input embedding}

Let $R$ be a real positive number. We assume that $R$ can be chosen
sufficiently large below. $R$ is important for implementing the closed
form solution based on \cite{Akyurek2022}.

In the input embedding, we set 
\begin{align}
\label{eq:Win}
\Win=\begin{bmatrix}
      1 & 0 & 0\\
0 & \frac{1}{RN} & 0\\
0 & 0 & \frac{1}{RN}\\
\sqrt{2} & -\frac{1}{RN} & -\frac{1}{RN}
     \end{bmatrix} 
\end{align}
and $\bin=\O_{4,1}$.  By (\ref{eq:input-embedding}), the $n$-th row of
the input embedding is
\begin{align}
\label{eq:Q_m_1}
\Q_{m,1}[n,:]^\top
=
\begin{bmatrix}
1\\
\frac{1}{RN}x_{m,n}\\
\frac{1}{RN}y_{m,n}\\
\sqrt{2}-\frac{1}{RN}x_{m,n}-\frac{1}{RN}y_{m,n}
\end{bmatrix},
\end{align}
which is the input to the transformer.

\subsection{Output of the first layer}

We here construct the first layer.

We consider the first head. We have
\begin{align}
\xi_{1,1,1}&:=(\M\Q_{m,1}[:,1]) ^\top(\M\Q_{m,1}[:,1])=N\\
\label{eq:xi_1_1_2}
\xi_{1,1,2}&:=(\M\Q_{m,1}[:,1])^\top(\M\Q_{m,1}[:,2])=\frac{1}{R}\ox_m\\
\xi_{1,1,3}&:=(\M\Q_{m,1}[:,1])^\top(\M\Q_{m,1}[:,3])=\frac{1}{R}\oy_m\\
\xi_{1,1,4}&:=(\M\Q_{m,1}[:,1])^\top(\M\Q_{m,1}[:,4])\notag\\
&=\sqrt{2}N
-\frac{1}{R}\ox_m-\frac{1}{R}\oy_m.
\end{align}
Therefore, we have
\begin{align}
(\M\Q_{m,1})^\top\M\Q_{m,1}=
\begin{bmatrix}
 \xi_{1,1,1} & \xi_{1,1,2} & \xi_{1,1,3} & \xi_{1,1,4}\\
 - & - & - & -\\
 - & - & - & -\\
 - & - & - & -\\
\end{bmatrix},
\end{align}
where the irrelevant elements are indicated by ``$-$''.
We set $\vTheta_{1,1}=\{\W_{1,1,1},\W_{1,1,2},\W_{1,1,3}\}$ in which
\begin{align}
\W_{1,1,1}&=\begin{bmatrix}
            1 & 0 & 0 & 0\\
            0 & 0 & 0 & 0\\
            0 & 0 & 0 & 0\\
            0 & 0 & 0 & 0\\
           \end{bmatrix}\\
\W_{1,1,2}&=\begin{bmatrix}
            1 & 0 & 0 & 0\\
            0 & -\frac{1}{\sqrt{2}} & 0 & 0\\
            0 & 0 & -\frac{1}{\sqrt{2}} & 0\\
            0 & 0 & 0 & \frac{-1}{\sqrt{2}}\\
           \end{bmatrix}\\
\W_{1,1,3}&=\begin{bmatrix}
            1 & 0 & 0 & 0\\
            0 & 0 & 0 & 0\\
            0 & 0 & 0 & 0\\
            0 & 0 & 0 & 0\\
           \end{bmatrix}.
\end{align}
Then, by (\ref{eq:LSA}) and (\ref{eq:LSAoutput}), the $n$-th row of the
output of the first head is
\begin{align}
\H_{m,1,1}[n,:]^\top
=\begin{bmatrix}
N\\  
-\frac{1}{\sqrt{2}R}\ox_m\\
-\frac{1}{\sqrt{2}R}\oy_m\\
-N+\frac{1}{\sqrt{2}R}\ox_m
+\frac{1}{\sqrt{2}R}\oy_m
 \end{bmatrix}.
\end{align}

On the other hand, for the second head, we set 
$\vTheta_{1,2}=\{\W_{1,2,1},\W_{1,2,2},\W_{1,2,3}\}$ in which
\begin{align}
\W_{1,2,1}=\W_{1,2,2}=\W_{1,2,3}=\O_{d,d}.
\end{align}
Then, by (\ref{eq:LSA}) and (\ref{eq:LSAoutput}), the $n$-th row of the
output of the second head is
\begin{align}
\H_{m,1,2}[n,:]=\O_{1,4}. 
\end{align}
Hence, there is no contribution of the second head.

By (\ref{eq:MHLSAoutput}), the $n$-th row of the output of
the MHLSA in the first layer is
\begin{align}
\label{eq:const-MHLSAoutput1}
\H_{m,1}[n,:]^\top&=\H_{m,1,1}[n,:]^\top\notag\\
&=
\begin{bmatrix}
N\\
-\frac{1}{\sqrt{2}R}\ox_m\\
-\frac{1}{\sqrt{2}R}\oy_m\\
-N+\frac{1}{\sqrt{2}R}\ox_m
+\frac{1}{\sqrt{2}R}\oy_m
\end{bmatrix}
\end{align}

Now, we consider the layer normalization. 
By (\ref{eq:LN-mu}), (\ref{eq:LN-sigma2}) and 
(\ref{eq:const-MHLSAoutput1}), for $\H_{m,1}[n,:]$, we have
\begin{align}
\label{eq:mu_m_1_n}
\mu_{m,1,n}&=0\\
\label{eq:sigma2_m_1_n}
\sigma^2_{m,1,n}&=\frac{1}{4}\left(
N^2+O\left(\frac{1}{R^2}\right)+N^2+O\left(\frac{N}{R}\right)\right)
\simeq \frac{N^2}{2},
\end{align}
since we can set a sufficiently large value for $R$; e.g.  $R=N^p$,
where $p$ is a sufficiently large positive integer. 

By (\ref{eq:after-LN}),
(\ref{eq:const-MHLSAoutput1}), 
(\ref{eq:mu_m_1_n}) and 
(\ref{eq:sigma2_m_1_n}), the $n$-th row of the output of the layer
normalization is
\begin{align}
\label{eq:oH_m_1}
\oH_{m,1}[n,:]^\top=
\begin{bmatrix}
 \sqrt{2}\\
-\frac{1}{RN}\ox_m\\
-\frac{1}{RN}\oy_m\\
-\sqrt{2}
+\frac{1}{RN}\ox_{m,n}
+\frac{1}{RN}\oy_{m,n}
\end{bmatrix}. 
\end{align}

Then, by (\ref{eq:after-skip-connection}), (\ref{eq:Q_m_1}) and
(\ref{eq:oH_m_1}), the $n$-th row of the output of the first 
layer is
\begin{align}
\label{eq:Q_m_2}
\Q_{m,2}[n,:]^\top=
\begin{bmatrix}
1+ \sqrt{2}\\
\frac{1}{RN}\Delta x_{m,n}\\
\frac{1}{RN}\Delta y_{m,n}\\
-\frac{1}{RN}(\Delta x_{m,n}+\Delta y_{m,n})\\
\end{bmatrix},
\end{align}
where
\begin{align}
\label{eq:Delta_x_m_n}
\Delta x_{m,n}&=x_{m,n}-\ox_m\\ 
\label{eq:Delta_y_m_n}
\Delta y_{m,n}&=y_{m,n}-\oy_m.
\end{align}

\subsection{Output of the second layer}

We next construct the second layer.

By (\ref{eq:Q_m_2}), we have
\begin{align}
\xi_{2,2,2}&:=(\M\Q_{m,2}[:,2])^\top (\M\Q_{m,2}[:,2])\notag\\
&=\frac{1}{R^2N}V_m\\
\xi_{2,2,3}&:=(\M\Q_{m,2}[:,2])^\top (\M\Q_{m,2}[:,3])\notag\\
&=\frac{1}{R^2N}C_m.
\end{align}
Thus, we have
\begin{align}
\label{eq:Q_m_2TQ_m_2}
(\M\Q_{m,2})^\top(\M\Q_{m,2})=\begin{bmatrix}
- & - & - & -\\
- & \xi_{2,2,2} & \xi_{2,2,3} & - \\
- & - & - & - &\\
- & - & - & - &\\
\end{bmatrix}.
\end{align}

We construct the first head.  To do this, we set
$\vTheta_{2,1}=\{\W_{2,1,1},\W_{2,1,2},\W_{2,1,3}\}$, in which
\begin{align}
\W_{2,1,1}&=
\begin{bmatrix}
0 & 1 & 0 & 0\\
0 & 0 & 0 & 0\\
0 & 0 & 0 & 0\\
0 & 0 & 0 & 0\\
\end{bmatrix}\\
\W_{2,1,2}&=
\begin{bmatrix}
0 & 0 & 0 & 0\\
0 & 0 & 0 & 0\\
0 & R^2N/\sqrt{2} & -R^2N/\sqrt{2} & 0\\
0 & 0 & 0 & 0\\
\end{bmatrix}\\
\W_{2,1,3}&=
\begin{bmatrix}
0 & 0 & 0 & 0\\
1 & 0 & 0 & 0\\
0 & 0 & 0 & 0\\
0 & 0 & 0 & 0\\
\end{bmatrix}.
\end{align}
Then, by (\ref{eq:LSA}), (\ref{eq:LSAoutput}) and
(\ref{eq:Q_m_2TQ_m_2}), the $n$-th row of the output of the first head
is
\begin{align}
\label{eq:H_m_2_1}
\H_{m,2,1}[n,:]^\top
=\begin{bmatrix}
0\\
\frac{1}{\sqrt{2}RN}C_m\Delta x_{m,n}\\
-\frac{1}{\sqrt{2}RN}C_m\Delta x_{m,n}\\
0\\
 \end{bmatrix}.
\end{align}

On the other hand, we consider the second head.
We set $\vTheta_{2,2}=\{\W_{2,2,1},\W_{2,2,2},\W_{2,2,3}\}$, in which
\begin{align}
\W_{2,2,1}&=
\begin{bmatrix}
0 & 1 & 0 & 0\\
0 & 0 & 0 & 0\\
0 & 0 & 0 & 0\\
0 & 0 & 0 & 0\\
\end{bmatrix}\\
\W_{2,2,2}&=
\begin{bmatrix}
0 & 0 & 0 & 0\\
R^2N & 0 & 0 & -R^2N\\
0 & 0 & 0 & 0\\
0 & 0 & 0 & 0\\
\end{bmatrix}\\
\W_{2,2,3}&=
\begin{bmatrix}
1/(1+ \sqrt{2}) & 0 & 0 & 0\\
0 & 0 & 0 & 0\\
0 & 0 & 0 & 0\\
\end{bmatrix}.
\end{align}
Then, by (\ref{eq:LSA}), (\ref{eq:LSAoutput}) and
(\ref{eq:Q_m_2TQ_m_2}), the $n$-th row of the output of the second head
is
\begin{align}
\label{eq:H_m_2_2}
\H_{m,2,2}[n,:]^\top
=\begin{bmatrix}
V_m\\
0\\
0\\
-V_m\\
 \end{bmatrix}.
\end{align}

Therefore, by (\ref{eq:MHLSAoutput}), 
(\ref{eq:H_m_2_1}) and (\ref{eq:H_m_2_2}), the $n$-th row of the output of
the MHLSA in the second layer is
\begin{align}
\label{eq:const-MHLSAoutput2}
\H_{m,2}[n,:]^\top
=\begin{bmatrix}
V_m\\
\frac{1}{\sqrt{2}RN}C_m\Delta x_{m,n}\\
-\frac{1}{\sqrt{2}RN}C_m\Delta x_{m,n}\\
-V_m\\
 \end{bmatrix}.
\end{align}

Now, we consider the layer normalization. 
By (\ref{eq:LN-mu}), (\ref{eq:LN-sigma2}) and 
(\ref{eq:const-MHLSAoutput2}), for $\H_{m,2}[n,:]$, we have
\begin{align}
\mu_{m,2,n}&=0\\
\label{eq:sigma2_m_2_n}
\sigma^2_{m,2,n}&=\frac{1}{4}\left(
2V_m^2+O\left(\frac{1}{R^2N^2}\right)\right)
\simeq \frac{1}{2}V_m^2
\end{align}
since $R$ is sufficiently large.
By (\ref{eq:after-LN}), the $n$-th row of the output of the layer normalization is
\begin{align}
\label{eq:oH_m_2}
\oH_{m,2}[n,:]^\top=
\begin{bmatrix}
\sqrt{2}\\
\frac{1}{RN}\frac{C_m}{V_m}\Delta x_{m,n}\\
-\frac{1}{RN}\frac{C_m}{V_m}\Delta x_{m,n}\\
-\sqrt{2}\\
\end{bmatrix}. 
\end{align}
The division performed by the layer normalization is actually used for
this purpose in \cite{Akyurek2022}.

Then, by (\ref{eq:after-skip-connection}), (\ref{eq:Q_m_2}) and
(\ref{eq:oH_m_2}), the $n$-th row of the output of the second layer is
\begin{align}
\label{eq:Q_m_3}
\Q_{m,3}[n,:]^\top=
\begin{bmatrix}
1+ 2\sqrt{2}\\
\frac{1}{RN}\Delta x_{m,n}+
\frac{1}{RN}\frac{C_m}{V_m}\Delta x_{m,n}\\
\frac{1}{RN}\Delta y_{m,n}
-\frac{1}{RN}\frac{C_m}{V_m}\Delta x_{m,n}\\
-\sqrt{2}
\end{bmatrix},
\end{align}
where we omit the $O\left(1/(RN)\right)$-term in $\Q_{m,3}[n,4]$.

\subsection{Model output}

Since $x_{m,N+1}=u_m$ and $y_{m,N+1}=0$, we have
\begin{align}
\label{eq:Q_m_3-N+1_3}
\Q_{m,3}[N+1,3]=-\frac{1}{RN}
\left(\oy_m+\frac{C_m}{V_m}(u_m-\ox_m)\right) 
\end{align}
by (\ref{eq:Q_m_3}).  
Now, $\Q_{m,3}$ is flattened into $\q_m$.
Then, we have $\q_m[D(N+1)-1]=\Q_{m,3}[N+1,3]$.
Hence, if we set
\begin{align}
\label{eq:Wout}
\Wout[i]=\begin{cases}
          -RN & i=D(N+1)-1\\
0 & {\rm otherwise}
         \end{cases}
\end{align}
then the transformer output for the $m$-th prompt is, approximately, 
\begin{align}
\label{eq:final-tf-output}
\TFout(\P_m)=\oy_m+\frac{C_m}{V_m}(u_m-\ox_m)
\end{align}
by (\ref{eq:transformer-output}) and (\ref{eq:Q_m_3-N+1_3}).  This is
consistent with the least squares estimate at $u_m$ in
(\ref{eq:lse-estimate-at-u_m}).

\subsection{Discussion}
\label{sec:discussion}

We have several remarks.

\begin{itemize}
 \item We introduced a partial masking to LSA.  This is because, for example as
in (\ref{eq:xi_1_1_2}), we need to eliminate the effect of the
prediction point $(u_m,0)$ on the calculation of the mean of the
in-context input samples $\ox_m$. In other words, it is necessary to treat the
in-context samples and prediction point separately here. This is not
possible since the attention mechanism calculates the inner product of
the two sequences, which implies that information on the sequence
dependent property disappears; e.g. see (\ref{eq:LSA}). In large
language models, this fact is generally important since we may need
masking and/or positional encoding to control the sequence dependent
property that cannot be controlled by the attention mechanism.

\item It is interesting that the features such as the mean of inputs and
       outputs for a simple regression problem are extracted in the first (shallow)
       layer and those are used to calculate
$\ey_m(u_m)$ in (\ref{eq:lse-estimate-at-u_m}) in the second (deep)
       layer. This is a natural arrangement in constructing $\ey_m(u_m)$
       and stems from computational limitations in an attention layer;
       i.e. the number of multiplications.

\item A large value for $R$ is significantly required for
performing division by layer normalization. Note that this enforces
the absolute values of weights to be either close to zero or extremely
large. 

\item Our construction may be relatively compact.
As shown in Appendix \ref{sec:appendix-nonuniqueness},
the representation of LSA is not unique in the sense
that, it produces the same output
for different inputs. Therefore, it might be possible to create the
same output using a more verbose expression, in which the most weight
values are not zeros. Actually, this will be seen in the numerical
       experiment.

\end{itemize}

\section{Numerical experiment}

We here show several numerical experiments to demonstrate that, given
the non-uniqueness of th expression noted in
Appendix \ref{sec:appendix-nonuniqueness}, our construction in the previous
section is actually used in the trained transformer.

\subsection{Setting of experiment}

The transformer defined in this paper is coded by PyTorch, in which we
omit the biases in the input and output linear transformation.  For the
training data $\D_m$ defined by (\ref{eq:D_m}), we set $\sigma^2=0.04$
for the noise variance and, at each $m$, the inputs including prediction
points are randomly drawn from a standard normal distribution
$N(0,1)$. And, we generate $\theta_{m,0}$ and $\theta_{m,1}$ according
to $N(0,1)$ independently at each $m$.  The length of the input sequence
is $N+1=51$ (the number of in-context samples is $N=50$). The target
output (teacher) for the transformer is the least squares estimate at
the prediction point.  This is a more direct instruction than the output
data at the prediction point. The transformer is $\TF(4,2,2)$.  For the
training, the number of training data is $M=5000$, the number of
validation data is $1000$, the batch size is $500$, the algorithm is
Adam\cite{Adam} with the learning rate of $0.005$, the maximum number of
training epochs is $20000$.  We apply the layer normalization without an
element-wise affine setting in {\tt torch.nn.LayerNorm}. As shown in
Appendix \ref{sec:appendix-nonuniqueness}, there is a lack of uniqueness
in the LSA representation. Therefore, we apply the $\ell_1$
regularization to the output weights, by which non-contributing weights
are set to zero and important weights may be highlighted.  The
regularization parameter here is $0.00001$. This may make one expect a
relatively compact expression as obtained in our construction.  We
employ the transformer that minimizes the validation error calculated
every $100$ epochs.

\subsection{Verifying the use of layer normalization}
\label{sec:prediction_correlation}

We execute $10$ different training runs under the above setting, in
which both of data and initialization differ for each run.  We generate
$1000$ new data of pairs of the prompt and the least squares estimate,
in which the prompts are i.i.d. samples from the same distribution as
the prompt in the training data. Then, for each prompt, we obtain the
output of the trained transformer and can calculate the squared test
error between the transformer output and the least squares estimate at
the prediction point. Thus, in each run, we can obtain the average test
error for $1000$ new data.

Simultaneously, for each $m$-th prompt in the new data, we can calculate
$\H_{m,2}$ defined in (\ref{eq:MHLSAoutput}).  The variance along the
embedding used in the layer normalization is the variance of
$\H_{m,2}[n,:]$ for the $n$-th sequence, which is $\sigma_{m,2,n}^2$ as
defined in (\ref{eq:LN-sigma2}). Also, we can calculate $V_m$ defined
in (\ref{eq:Vm}) as the variance of in-context input samples at each $m$
in the new data.  In our construction, the division by $V_m$ in
(\ref{eq:lse-estimate-at-u_m}) is executed using the division by the
standard deviation in the layer normalization in (\ref{eq:oH_m_2}).  To
see this, we calculate the correlation coefficient between
$1/\sigma_{m,2,51}$ and $1/V_m$ for $m=1,\ldots,1000$ in each run.

The results for $10$ runs are summarized in Table
\ref{tbl:error-correlation}. We can see that the correlation coefficient
is larger than $0.9$ in the first and fifth runs. This implies that the
division by $V_m$ is possible to be calculated using the layer
normalization for these two runs. Additionally, the trained transformer
in the first run shows a better predictive performance in terms of the
average test error. Then, we further analyze the trained transformer in
the first run below. 

\begin{table}[ht]
\begin{center}
\caption{Average squared test error and correlation between reciprocals
of standard deviation in layer normalization and variance of in-context input samples}
\label{tbl:error-correlation}
\begin{tabular}{|c|c|c|}\hline
No. & Average test error & Correlation\\\hline
1 & 0.000058  &  0.966911 \\\hline
2 & 0.000269  &  0.417908 \\\hline
3 & 0.000307  &  0.535359 \\\hline
4 & 0.000101  &  0.662808 \\\hline
5 & 0.000109  &  0.988581 \\\hline
6 & 0.000095  &  0.354986 \\\hline
7 & 0.000096  &  0.816449 \\\hline
8 & 0.000114  &  0.802375 \\\hline
9 & 0.003176  &  0.223738 \\\hline
10 & 0.000090  &  0.457691 \\\hline
\end{tabular}
\end{center}
\end{table}

\subsection{Output weights}

\begin{figure}[h]
\centering
\includegraphics[width=80mm]{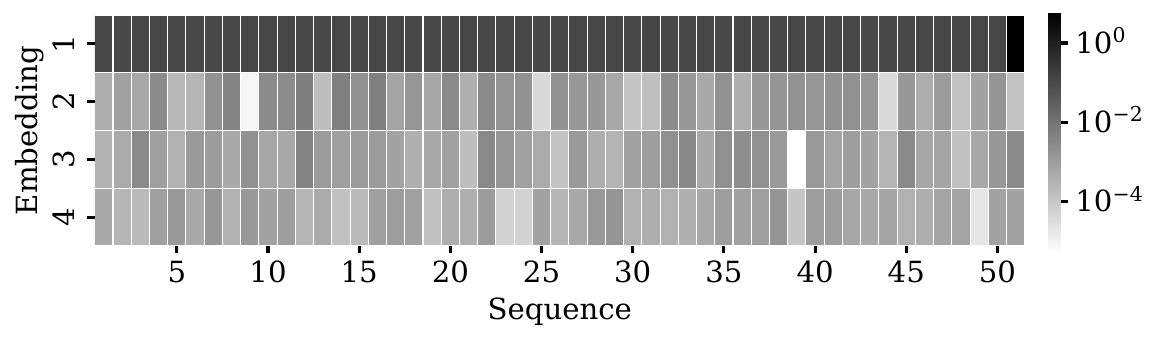}
\caption{Absolute output weight values}
\label{fig:weight_value}
\end{figure}

In Fig. \ref{fig:weight_value}, for the first run in Table
\ref{tbl:error-correlation}, we show the heat map of the absolute values
of the output weights which are denoted by $\Wout$ in this paper. The
absolute weight values are shown on a logarithmic scale. Although
$\Wout$ is a $204$-dimensional vector in our modeling, we here
convert it into a $4\times 51$ matrix whose row and column correspond to
the indices of embeddings and sequences respectively. The converted
$4\times 51$ output weight matrix is referred to as $\oWout$. We refer to
the final-layer-cell whose corresponding output weight is $\oWout[d,n]$
as the $[d,n]$ cell, which is the cell at the $d$-th embedding and the $n$-th sequence.

In Fig. \ref{fig:weight_value}, we can see that the absolute weight
values of the first embedding cells are relatively large and,
especially, the absolute value of $\oWout[1,51]$ is extremely large
compared to the other values. Therefore, the $[1,51]$ cell is the
dominant factor for the transformer output. Note that the contribution
of the first embedding  are larger than the contributions of the other
embeddings overall. 

\subsection{Fitting curve of in-context samples after training}

Here, for the first run, we show the fitting curves produced by the
trained transformer to the in-context samples.

We consider the prompt defined in (\ref{eq:slr-prompt}), in which the
in-context samples are $\D_m$ and the prediction point is $u_m=u$.  If
we keep $\D_m$ the same and change only $u$, the transformer is a
function of $u$.  We define the output of the
$[d,n]$ cell in the final layer for $u$ under a fixed $\D_m$ by
$q_{d,n}(u|\D_m)$.  Then, the transformer output for $u$ is given by
\begin{align}
t(u|\D_m)&:=\sum_{d=1}^{4}\sum_{n=1}^{51}\oWout[d,n]q_{d,n}(u|\D_m).
\end{align}
We also define
\begin{align}
t_0(u|\D_m)&:=\oWout[1,51]q_{1,51}(u|\D_m)\\
t_1(u|\D_m)&:=\sum_{n=1}^{50}\oWout[1,n]q_{1,n}(u|\D_m).
\end{align}
The former is the contribution of the $[1,51]$ cell and the latter
is the contribution of the other cells in the first embedding.
Then, we generate $100$ new data for the prompts to the trained
transformer, which are $\D_{50}=\{(x_1,y_1),\ldots,(x_n,y_{50}),(u_k,0)\}$ for
$k=1,\ldots,100$, where $(x_i,y_i)$ is generated from the same
distribution as the training data and $u_k$ is the equidistant points in
$[-4,4]$.  In other words, we obtain the transformer output at $100$
different prediction points under the same in-context samples.
Therefore, we can plot a fitting curve for the in-context samples.

\begin{figure}[p]
  \centering
\begin{minipage}{40mm}
\centering
  \includegraphics[width=40mm]{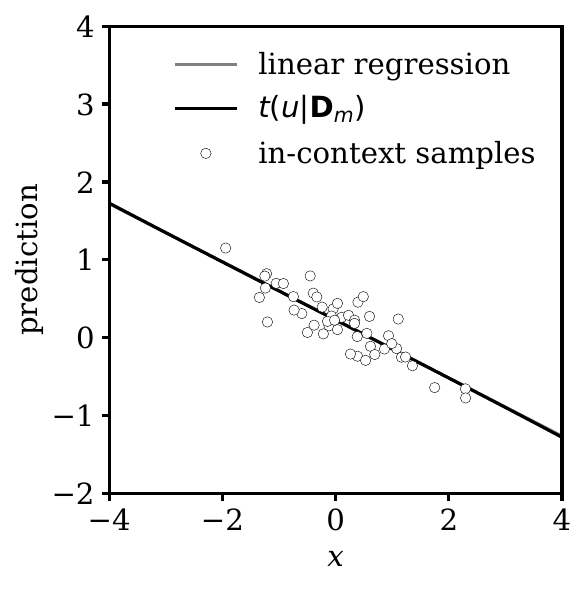}\\
(a) $t(u_k|m)$ in Example A.
\end{minipage}
\begin{minipage}{40mm}
\centering
  \includegraphics[width=40mm]{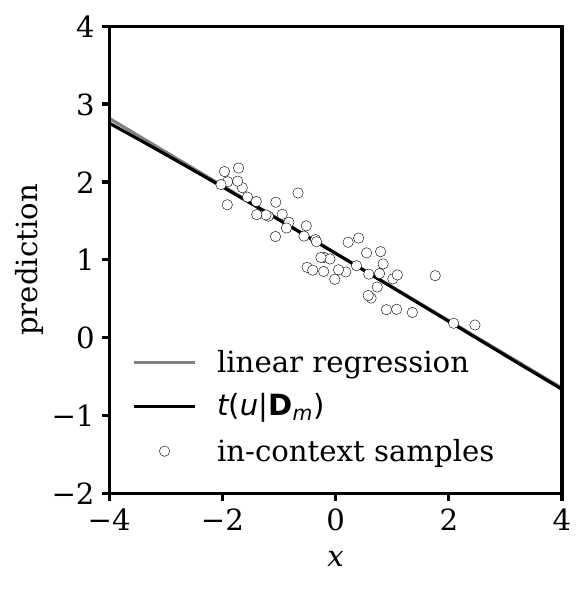}\\
(b) $t(u_k|m)$ in Example B.
\end{minipage}

\vspace{2mm}

\begin{minipage}{40mm}
\centering
  \includegraphics[width=40mm]{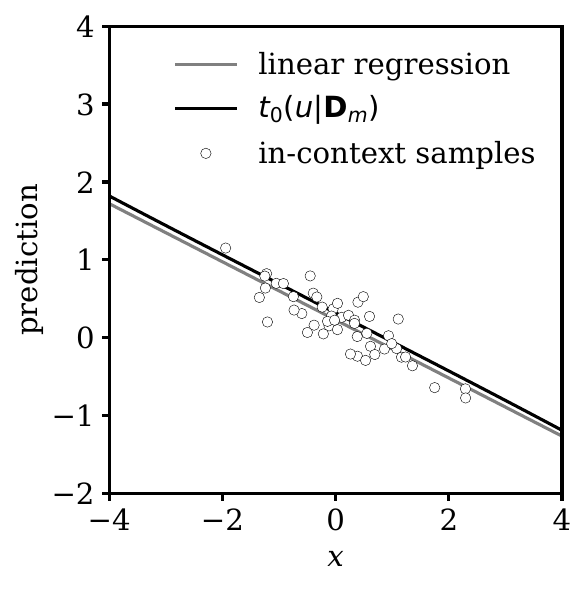}\\
(c) $t_0(u_k|m)$ in Example A.
\end{minipage}
\begin{minipage}{40mm}
\centering
  \includegraphics[width=40mm]{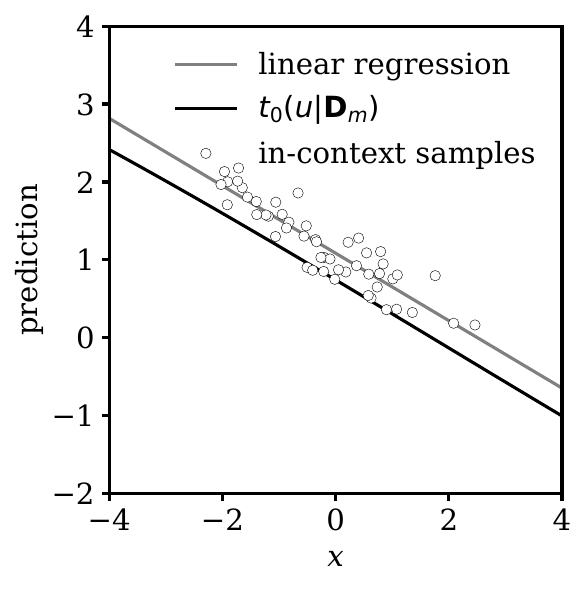}\\
(d) $t_0(u_k|m)$ in Example B.
\end{minipage}

\vspace{2mm}

\begin{minipage}{40mm}
\centering
  \includegraphics[width=40mm]{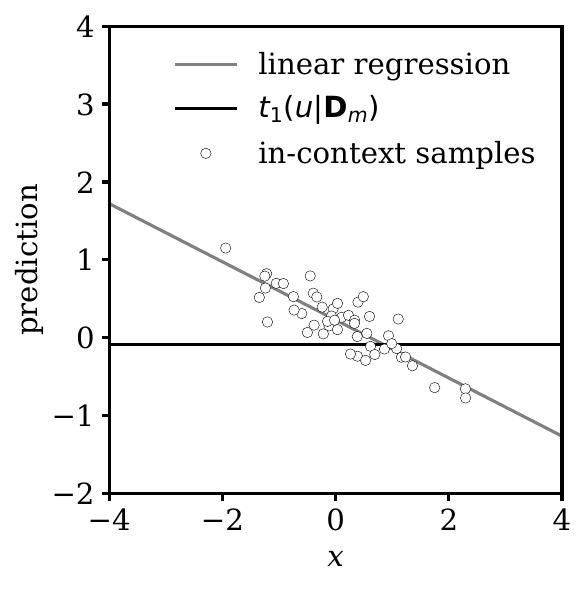}\\
(e) $t_1(u_k|m)$ in Example A.
\end{minipage}
\begin{minipage}{40mm}
\centering
  \includegraphics[width=40mm]{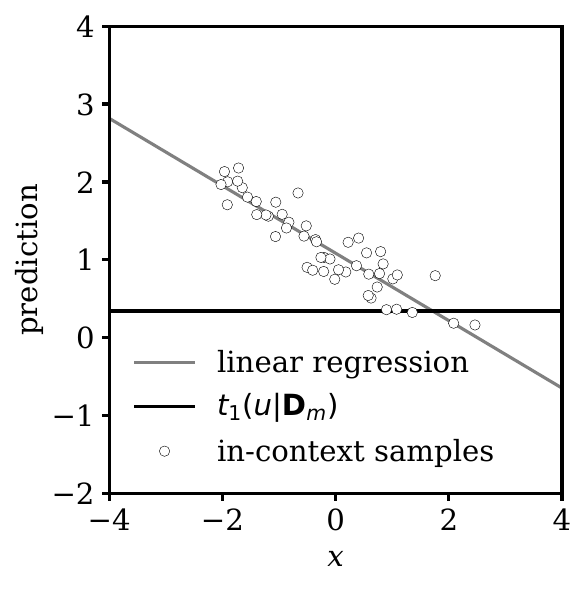}\\
(f) $t_1(u_k|m)$ in Example B.
\end{minipage}

 \caption{Examples of fitting curves after training}
  \label{fig:fitting_example}
\end{figure}

In Fig. \ref{fig:fitting_example}, we show two examples, denoted by A
and B, of in-context samples (open circle) and graphs of
$t(u_k|\D_{50})$, $t_1(u_k|\D_{50})$, $t_0(u_k|\D_{50})$ (black solid
line), in which (a), (c) and (e) for Example A and (b), (d), (f) for
Example B. In this figure, we also show the graph of the least squares
estimate of a simple regression model (gray solid line), which is
obtained for the in-context samples, $\D_{50}$.  Note that the graph of
the least squares estimate is a straight line in a simple regression
problem. From (a) and (b), we can see that the transformer output,
$t(u_k|\D_{50})$, is well consistent with the least squares estimate. In
(c) and (d), the contributions through the $[1,51]$ cells,
$t_0(u_k|\D_{50})$, slightly deviate from the least squares estimate,
while the slopes of both are well consistent with the slopes of the
least squares estimates. In (e) and (f), the contributions trough the
other cells in the first embedding, $t_1(u_k|\D_{50})$, are constant for
both.  Actually, the variance of $t_1(u_k|\D_{50})$, $k=1,\ldots,100$
are $2.220446\times 10^{16}$ for Example A and $8.881784\times10^{-16}$
for Example B, which implies that there are no changes in the
outputs. Therefore, $t_0(u_k|\D_{50})$ captures the slope of the the
least squares estimates precisely and $t_1(u_k|\D_{50})$ is responsible
for correcting the constants.  The average test error using only the
first embedding was $0.000056$, which is almost consistent with the
error given by the transformer output of the first run in Table
\ref{tbl:error-correlation}.  This implies that the transformer output
is almost determined by the contribution from the first embedding;
i.e. $t_0(u_m|\D_m)+t_1(u_m|\D_m)$.  Note that, indeed, it is a little
small since the contribution from the other embedding may cause problems
near the edges of the input range, in which there are few training
samples for transformers. In the next section, using regression
analysis, we clarify how the trained transformer implements these
predictions.

\subsection{Regression analysis}

For the first run in the numerical experiment in Section
\ref{sec:prediction_correlation}, we can obtain $t(u_m|\D_m)$,
$t_0(u_m|\D_m)$ and $t_1(u_m|\D_m)$ for $m=1,\ldots,1000$, which are the
final and intermediate outputs in the trained transformer.  Then, we
consider linear regression analysis of these targets by
\begin{align}
\label{eq:regression_analysis}
f_m:=&\alpha_0+\alpha_1u_m+\alpha_2\ox_m+
\alpha_3\oy_m+\alpha_4\frac{C_m}{V_m}u_m
+\alpha_5\frac{C_m}{V_m}\ox_m\notag\\ 
&+\alpha_6\frac{\ox_m}{V_m}
+\alpha_7\frac{\oy_m}{V_m}
+\alpha_8\frac{1}{V_m}+\alpha_9\frac{C_m}{V_m},
\end{align}
where $u_m,\ox_m,\oy_m,\ldots,C_m/V_m$ are explanatory variables (features)
and $\alpha_k$, $k=0,1,\ldots,9$ are coefficients.  This is viewed as a
linear probe in \cite{Akyurek2022}. The explanatory variables are
selected to include the terms in the solution
(\ref{eq:lse-estimate-at-u_m}) and to heuristically bring the multiple
correlation coefficient between $t_1(u_m|\D_m)$ and $f_m$ closer to $1$.
In Table \ref{tbl:regression_analysis}, we show the $R^2$ score and the
estimated coefficients.
We summarize the results below.

\begin{table}[ht]
\begin{center}
\caption{$R^2$ and estimated coefficients in regression analysis of 
$t(u_m|\D_m)$, $t_0(u_m|\D_m)$ and $t_1(u_m|\D_m)$.}
\label{tbl:regression_analysis}
\begin{tabular}{|c|c|c|c|}\hline
 & $t(u_m|\D_m)$ & $t_0(u_m|\D_m)$ & $t_1(u_m|\D_m)$\\\hline
$R^2$ & 0.999974 & 0.997697 & 0.957023\\\hline
$\alpha_0$ &  -0.00205  &  -0.018546  &  0.018853\\\hline
$\alpha_1$ & -0.000005 & 0.000670 & -0.001094\\\hline
$\alpha_2$ & -0.016581 & 0.374289 & -0.387986\\\hline
$\alpha_3$ & 1.000339 & 0.046484 & 0.951257\\\hline
$\alpha_4$ & 1.000394 & 1.003497 & -0.003356\\\hline
$\alpha_5$ & -1.002698 & -0.243266 & -0.757685\\\hline
$\alpha_6$ & 0.017339 & -0.251728 & 0.264673\\\hline
$\alpha_7$ & -0.000487 & 0.721800 & -0.720420\\\hline
$\alpha_8$ & 0.001127 & 0.193471 & -0.193076\\\hline
$\alpha_9$ & -0.000034 & 0.134298 & -0.133731\\\hline
\end{tabular}
\end{center}
\end{table}

\subsubsection{Fitting accuracy}

We can see that the $R^2$ scores are high for all cases. This implies
that the fittings by the linear regression models are valid. Note that
we do not need to worry about the effects of overfitting since the
number of samples is $1000$, which may be large enough for $10$
coefficients.

\subsubsection{Result for $t(u_m|\D_m)$}

For $t(u_m|\D_m)$, from the $R^2$ score, the fitting is almost
perfect. Looking at the estimated coefficients,
$(\alpha_3,\alpha_4,\alpha_5)\simeq (1,1,-1)$ and $\alpha_k\simeq 0$ for
the other terms.  These imply that the output of the trained transformer
is of the form of the least squares solution given in
(\ref{eq:lse-estimate-at-u_m}).

\subsubsection{Results for  $t_0(u_m|\D_m)$ and $t_1(u_m|\D_m)$}

At each explanatory variable, the sum of the estimated coefficients for
$t_0(u_m|\D_m)$ and $t_1(u_m|\D_m)$ is almost consistent with the
estimated coefficient for $t(u_m|\D_m)$ although, of course, it is not
completely consistent. The slope of transformer output is determined by
the term with $\alpha_4$, which is $u_mC_m/V_m$, and is almost
represented by $t_0(u_m|\D_m)$. On the other hand, the constant term of
the transformer output is determined by the terms with $\alpha_3$ and
$\alpha_5$, which are $\oy_m$ and $\ox_mC_m/V_m$.  The term with
$\alpha_3$ is almost represented by $t_1(u_m|\D_m)$ and the term with
$\alpha_5$ is represented by both of $t_0(u_m|\D_m)$ and
$t_1(u_m|\D_m)$. In the other term which is not related to the solution
given by (\ref{eq:lse-estimate-at-u_m}) , the coefficients for
$t_0(u_m|\D_m)$ and $t_1(u_m|\D_m)$ have almost the same magnitude but
opposite signs.  Therefore, those terms vanish in the transformer
output. 

\subsubsection{Relation to construction of transformer}

We explain the correspondence between our construction of a transformer
and the numerical example. The output of the $[1,n]$ cell in the
numerical example corresponds to $\Q_{m,3}[n,3]$ in (\ref{eq:Q_m_3}) in
our construction.

First, $\oy_m$, $C_mu_m/V_m$ and $C_m\ox_m/V_m$ are the terms
correspond to $\alpha_3$, $\alpha_4$ and $\alpha_5$ in Table
\ref{tbl:regression_analysis}
 respectively and are the components of the
least squares solution.  As mentioned above, the other terms in
$t_0(u_m|\D_m)$ and $t_1(u_m|\D_m)$ cancel each other.  If the
coefficients are zero for these terms for both of $t_0(u_m|\D_m)$ and
$t_1(u_m|\D_m)$, then the result remains the same.  Actually, if the
trained transformer follows our implementation, then this is achieved.  To
achieve this, the values of many of the weights in the self-attention
are required to be zero as in our construction. Of course, it may not
actually happen in training. Nevertheless, the output of the trained
transformer is consistent with the output of the transformer that we
construct. This is because of a lack of uniqueness of the representation
in a transformer as mentioned in Section \ref{sec:discussion}.

Now, we here focus on the main terms in the least squares solution.  By
(\ref{eq:Q_m_3}), $\Q_{m,3}[51,3]$ consists of the sum of $\oy_m$,
$C_mu_m/V_m$ and $-C_m\ox_m/V_m$ in (\ref{eq:Q_m_3}) since $x_{m,N+1}=0$
and $y_{m,N+1}=0$. Especially, as in our construction, only
$\Q_{m,3}[51,3]$ is responsible for $u_m$ that determines the slope of
the least squares estimate.  Actually, in Table
\ref{tbl:regression_analysis}, $\alpha_4$ is very small for
$t_1(u_m|\D_m)$ and is close to $1$ for $t_0(u_m|\D_m)$. However,
$\alpha_3$ and $\alpha_5$ are small in $t_1(u_m|\D_m)$ and those are
supplemented by $t_0(u_m|\D_m)$. So, the question is how the terms
$\alpha_3$ and $\alpha_5$ in $t_1(u_m|\D_m)$ are produced by
$\Q_{m,3}[n,3]$ for $n=1,\ldots,50$. As seen in
Fig. \ref{fig:weight_value}, the output weight values are almost
identical for $n=1,\ldots,50$ in the first embedding. Therefore, in our
construction, the weighted sum of the right hand side of
(\ref{eq:Q_m_3}) for $n=1,\ldots,50$ just consists of the weighted sum
of $\oy_m$ and $C_m\ox_m/V_m$ since the sums of $x_{m,n}$'s and
$y_{m,n}$'s reduce to the constant multiples of $\ox_m$ and $\oy_m$
respectively. This brings us the appropriate values for $\alpha_3$ and
$\alpha_5$ in $t_1(u_m|\D_m)$ to compensate what $t_0(u_m|\D_m)$ lacks,
which is a constant term in the fitting curve. This yields the results
for $\alpha_3$ and $\alpha_5$ in Table \ref{tbl:regression_analysis}.
After all, the least squares solution is obtained as the trained
transformer output under our construction of $\Q_{m,3}[n,3]$ in
(\ref{eq:Q_m_3}) while the usage of $\Q_{m,3}[n,3]$ in training does not
follow our construction; i.e. $(\alpha_3,\alpha_4,\alpha_5)=(1,1,-1)$
and the other coefficients are zero for $t_0(u_m|\D_m)$ and all the
coefficients are zero for $t_1(u_m|\D_m)$. This is again due to the
non-uniqueness of the LSA representation. Therefore, our construction
is valid except for the factor that comes from the non-uniqueness.

\section{Conclusion and future work}

In this paper, we constructed the transformer that outputs the least
squares estimate of a simple regression problem, in which the
transformer comprises a linear transformation of the input, a stack of
transformer blocks, a flatten layer, and an output linear
transformation, in which the transformer block consists of multi-head
masked linear self-attention blocks with layer normalization followed by
a skip connection. To achieve this by $\TF(4,4,2)$, in the construction,
we represent a closed form solution by using the layer normalization,
which is an insight of \cite{Akyurek2022,Hagiwara2026}.  For in-context
learning of regression tasks, most studies have considered
implementations of the gradient descent algorithms. This is because
multiplication, which is a basic operation of the attention mechanism,
is suitable for implementing the gradient descent algorithm.  However,
the experimental evidence that the gradient descent algorithm is
implemented through training is still unclear. In contrast, in this
paper, we showed an experimental example, in which, by taking the
non-uniqueness of the transformer representation into account, our
implementation for a closed form solution is used in the transformer
trained with the $\ell_1$ regularization when the target output is the
least squares estimate.

We have several things to address along this line.
\begin{itemize}
\item We applied the $\ell_1$ regularization to eliminate the effects of
unnecessary weights. It seems that this helped concentrate the essential
output into a single embedding.  However, it does not seem natural and
may causes a bias in the prediction error. Therefore, we are interested
in the implementation under the training without $\ell_1$
regularization.

\item As shown in Appendix \ref{sec:appendix-gd}, the gradient descent based
implementation may not be comparable in $\TF(4,2,2)$ since the number of
layers may not be enough to implement the sufficient number of gradient
descent steps; e.g. see also \cite{Hagiwara2026}.  To put it another
way, the transformer obtain a closed form solution due to the limited
size. The gradient descent based implementation may be effective when
the size increases. The investigation of the impact of the size on the
implementation is left as future works.

\item The construction under the other prompt format should be examined.

\item Although, in our experiment, we employed the least squares estimate as
the target output in training, the output sample at the prediction point
may be natural as the target output. The theoretical and experimental
      analysis of this case is also left as a future work.

\end{itemize}



\appendix

\section*{Appendix}

\section{An example of a lack of uniqueness}
\label{sec:appendix-nonuniqueness}

We repost here a simple example of \cite{Hagiwara2026}, in which LSA
produces the same output for the different inputs. Here, we consider LSA
without a mask below for simplicity.

Let $\Q$ be an $N\times 3$ matrix whose $n$-th row is $[1,a_n,b_n]$
and define $a:=\sum_{n=1}^Na_n$ and $b:=\sum_{n=1}^Nb_n$.
In (\ref{eq:LSA}), we set this $\Q$ as the input matrix and we set
$\vTheta=\{\W_1,\W_2,\W_3\}$, where
\begin{align}
\W_1=\W_3=\begin{bmatrix}
      1 & 0 & 0\\
      0 & 0 & 0\\
      0 & 0 & 0\\
     \end{bmatrix},~
\W_2=\begin{bmatrix}
      0 & 0 & 0\\
      0 & 1 & 0\\
      0 & 0 & 1\\
     \end{bmatrix}.
\end{align}
Then, the $n$-th row of $\LSA_{\vTheta}(\Q)$ is $[0,a,b]$ for any $n$.
On the other hand, let $\Q$ be an $N\times 3$ matrix whose $n$-th row is
$[1,a_n,a_n+b_n]$. In (\ref{eq:LSA}), by setting
\begin{align}
\W_1=\W_3=\begin{bmatrix}
      1 & 0 & 0\\
      0 & 0 & 0\\
      0 & 0 & 0\\
     \end{bmatrix},~
\W_2=\begin{bmatrix}
      0 & 0 & 0\\
      0 & 1 & -1\\
      0 & 0 & 1\\
     \end{bmatrix},
\end{align}
the $n$-th row of $\LSA_{\vTheta}(\Q)$ is $[0,a,b]$ for any $n$.
This is the same as in the first case.

\section{Comparison to gradient descent}
\label{sec:appendix-gd}

We compare the trained transformer and the gradient descent method for a
simple regression problem.  Although a simple regression problem has two
parameters which are intercept and slope, we employ the one-parameter
version of the gradient descent, which is analytically derived.

We define
\begin{align}
f_{\theta_m}(u_m):=\oy_m+\theta_m(u_m-\ox_m), 
\end{align}
where $\theta_m$ is a parameter to be adjusted.  By
(\ref{eq:lse-estimate-at-u_m}), $\etheta_m=C_m/V_m$ is a closed
form solution to $\theta_m$.  We also define
\begin{align}
S_m(\theta):=\frac{1}{2}\sum_{n=1}^N(y_{m,n}-f_{\theta}(x_{m,n}))^2. 
\end{align}
We have
\begin{align}
\frac{\partial S_m(\theta)}{\partial\theta}
&=-\sum_{n=1}^N(y_{m,n}-f_{\theta}(x_{m,n}))(x_{m,n}-\ox_m)\notag\\
&=-N(C_m-\theta V_m).
\end{align}
By setting $\theta_m(0)$ as an initial value, at the $t$-th step,  
the update equation of the gradient descent for $\theta_m(t)$ is given by
\begin{align}
\theta_m(t)&=\theta_m(t-1)-\eta
\left.\frac{\partial S_m(\theta)}{\partial\theta}\right|_{\theta=\theta_m(t-1)}\notag\\
&=\theta_m(t-1)+N\eta(C_m-\theta_m(t-1)V_m),
\end{align}
where $\eta>0$ is a learning rate. The maximum number of repetitions is
denoted by $T$.

For the first run in the numerical experiment in Section
\ref{sec:prediction_correlation}, at each $m$, we obtain the linear
estimate $f_{\theta_m}$ using the gradient descent method based on the
in-context samples, where we set $\theta_m(0)=0$.  In Table
\ref{tbl:gd-testerror}, we show the average squared test error of the
gradient descent method for different $\eta$ and $T$.

\begin{table}[h]
\begin{center}
\caption{Average squared test error of gradient descent}
\label{tbl:gd-testerror}
\begin{tabular}{|c|c|c|}\hline
$\eta$ & test error($T=1$) & test error($T=4$)\\\hline
0.4 & 0.080785 & 0.000425\\\hline
0.45 & 0.050393 & 0.000130\\\hline
0.5 & 0.042642 & 0.000061\\\hline
0.55 & 0.057532 & 0.000309\\\hline
0.6 & 0.095063 & 0.002125\\\hline
\end{tabular}
\end{center}
\end{table}

In Table \ref{tbl:gd-testerror}, both of $T=1$ and $T=4$, $\eta=0.5$ is
the best choice for the leaning rate.  Since the average test error for
the trained transformer is $0.000058$ in Table
\ref{tbl:error-correlation}, the gradient descent method with $T=4$ is
comparable to the trained transformer by Table \ref{tbl:gd-testerror} if we
can choose $\eta=0.5$.  In other words, the gradient descent method
needs four steps to show the performance that is comparable to the
implementation of a closed form solution here.  For the prompt employed
in this paper, \cite{Hagiwara2026} showed that $\TF(4,3,2)$ can
implement the two-step gradient descent, in which the learning
rate is determined by the weight values of the transformer. Therefore,
an optimal learning rate may be obtained by optimizing the weight
values.  However, by the construction in \cite{Hagiwara2026},
$\TF(4,2,2)$ seems to be too simple to implement the four-step gradient
descent.

\end{document}